\documentclass[final]{cvpr}

\usepackage{times}
\usepackage{epsfig}
\usepackage{graphicx}
\usepackage{amsmath}
\usepackage{amssymb}
\usepackage{multirow}
\usepackage{makecell}
\usepackage{booktabs}
\usepackage{threeparttable}
\usepackage{arydshln} 
\usepackage{url}

\usepackage[pagebackref=true,breaklinks=true,colorlinks,bookmarks=false]{hyperref}

\def\cvprPaperID{519} 
\def\confYear{CVPR 2022}

\begin{document}

\title{Event-based Video Reconstruction via Potential-assisted Spiking Neural Network}

\author{Lin Zhu\textsuperscript{1,2}, Xiao Wang\textsuperscript{2}, Yi Chang\textsuperscript{2}, Jianing Li\textsuperscript{1,2}, Tiejun Huang\textsuperscript{1}, Yonghong Tian\textsuperscript{1,2,}\thanks{Corresponding author.}\\
Peking University\textsuperscript{1}, Peng Cheng Laboratory\textsuperscript{2}\\
}

\maketitle

\pagestyle{empty}
\thispagestyle{empty}
\begin{abstract}
Neuromorphic vision sensor is a new bio-inspired imaging paradigm that reports asynchronous, continuously per-pixel brightness changes called `events' with high temporal resolution and high dynamic range.
So far, the event-based image reconstruction methods are based on artificial neural networks (ANN) or hand-crafted spatiotemporal smoothing techniques.
In this paper, we first implement the image reconstruction work via deep spiking neural network (SNN) architecture.
As the bio-inspired neural networks, SNNs operating with asynchronous binary spikes distributed over time, can potentially lead to greater computational efficiency on event-driven hardware.
We propose a novel Event-based Video reconstruction framework based on a fully Spiking Neural Network (EVSNN), which utilizes Leaky-Integrate-and-Fire (LIF) neuron and Membrane Potential (MP) neuron. 
We find that the spiking neurons have the potential to store useful temporal information (memory) to complete such time-dependent tasks.
Furthermore, to better utilize the temporal information, we propose a hybrid potential-assisted framework (PA-EVSNN) using the membrane potential of spiking neuron. The proposed neuron is referred as Adaptive Membrane Potential (AMP) neuron, which adaptively updates the membrane potential according to the input spikes.
The experimental results demonstrate that our models achieve comparable performance to ANN-based models on IJRR, MVSEC, and HQF datasets.
The energy consumptions of EVSNN and PA-EVSNN are 19.36$\times$ and 7.75$\times$ more computationally efficient than their ANN architectures, respectively. The code and pretrained model are available at \url{https://sites.google.com/view/evsnn}.
\end{abstract}

\section{Introduction}
%
Event cameras~\cite{1,2} are bio-inspired vision sensors that pose a paradigm shift in the way visual information is acquired.
Compared with standard cameras, event cameras have high temporal resolution, high dynamic range (140 dB vs. 60 dB of standard cameras), and low power consumption.
Event cameras work asynchronously, recording the stream of events ($t,x,y,p$) which includes the timestamp, pixel location and polarity of the brightness changes.

Despite the advantages of the event data, it is not friendly to human vision and traditional computer vision~\cite{scheerlinck2020fast,wang2021visevent}.
As a solution, image reconstruction bridges the gap between human visualization and events, giving us an intuition of the rich information encoded by events.
On other hand, image is a useful representation for conventional frame-based computer vision~\cite{rebecq2019events}.
Reconstructing images from asynchronous events has been explored in various researches. 
Early works attempt to recover the intensity of an image from events based on hand-crafted priors~\cite{scheerlinck2018continuous, scheerlinck2019asynchronous, bardow2016simultaneous,munda2018real}.
Recently, deep neural network based reconstruction models~\cite{wang2019event,rebecq2019events, rebecq2019high,scheerlinck2020fast,stoffregen2020reducing,paredes2021back,weng2021event} have demonstrated impressive performance.
The events usually be transformed in to time-surfaces, event images or voxel grids as the input of convolutional neural network.
However, large artificial neural networks (ANN) can be memory and computationally intensive~\cite{scheerlinck2020fast}, consuming power and hampering the low latency of event cameras.

In fact, the sparse event data can be effectively combined with neuromorphic hardware for low-power spiking neural network (SNN) applications~\cite{haessig2018spiking}.
Compared with ANN, SNN is more biologically realistic and its neurons communicate with each other via discrete spikes instead of continuous-valued activations. 
Visual systems~\cite{orchard2015hfirst, amir2017low} constructed with SNN and event cameras have demonstrated their capacity in solving visual tasks as well as prominent energy-efficiency.
However, most of the SNN work has so far been focused on problems like classification~\cite{fang2021incorporating, xing2020new, zheng2020going}, optical estimation~\cite{paredes2019unsupervised,hagenaars2021self}, motion segmentation~\cite{parameshwara2021spikems}, and angular velocity regression~\cite{gehrig2020event}.
To the best of our knowledge, we are the first to attempt image reconstruction task based on a deep SNN architecture.

In this paper, we propose a novel Event-based Video reconstruction framework based on a fully Spiking Neural Network (EVSNN), which utilizes the Leaky-Integrate-and-Fire (LIF) neurons and a Membrane Potential (MP) neuron. To better extract the temporal information, we propose a hybrid potential-assisted framework (PA-EVSNN) using the membrane potential of spiking neurons.
The main contributions of this paper are summarized as follows:

1) We first explore a fully spiking neural network (EVSNN) architecture on event-based image reconstruction, which utilizes LIF neuron and MP neuron. This is also the first attempt to develop a deep SNN for image reconstruction task.

2) We propose a hybrid potential-assisted SNN (PA-EVSNN), which uses adaptive membrane potential (AMP) neurons to improve the temporal receptive field of EVSNN. AMP neurons can adjust the membrane time constant according to the input spike to adapt to various reconstruction scenes.

3) The experiments on public datasets demonstrate that the proposed models have comparable performance to existing ANN-based models, while the energy consumptions of EVSNN and PA-EVSNN are 19.36$\times$ and 7.75$\times$ more computationally efficient than their ANN architectures, respectively. Compared to E2VID, the proposed EVSNN and PA-EVSNN achieve 24.15$\times$ and 8.76$\times$ more computationally efficient improvement, respectively.

\section{Related work}
\noindent\textbf{Spiking Neural Network}
Supervised learning of SNNs was first proposed by SpikeProp~\cite{bohte2002error}, it used a linear approximation to overcome the non-differentiable threshold-triggered firing mechanism of SNNs, backpropagation was utilized to update weight.
Some works applied to single-layer SNN optimization appeared, including Tempotron~\cite{gutig2006tempotron}, Re-SuMe~\cite{ponulak2010supervised}, and SPAN~\cite{mohemmed2012span}. 
Recently, the surrogate gradient method provides an effective solution for training multi-layer SNN~\cite{lee2016training, jin2018hybrid, wu2018spatio, shrestha2018slayer, lee2020enabling, kaiser2020synaptic}.
It utilized surrogate derivatives to define the derivative of the threshold-triggered firing mechanism. 
Therefore, SNN can be optimized by gradient descent algorithm like ANNs, which makes the training of deep SNN possible.



Most of the learning-based SNN work has so far been focused on problems like classification~\cite{fang2021incorporating, xing2020new, zheng2020going}, optical estimation~\cite{paredes2019unsupervised,hagenaars2021self,hagenaars2021self}, motion segmentation~\cite{parameshwara2021spikems}, and angular velocity regression~\cite{gehrig2020event}.
There are also some unsupervised SNNs~\cite{zhu2020retina,zheng2021high} proposed for image reconstruction based on spike camera~\cite{zhu2019retina}.
Among them, ~\cite{hagenaars2021self} and ~\cite{ranccon2021stereospike} utilized deep SNNs for optical and depth estimation, respectively. 
In addition, Lee \emph{et al.}~\cite{lee2020spike} proposed an ANN-SNN hybrid architecture for optical estimation, using SNN as encoder and ANN as decoder and residual block. Zhang \emph{et al.}~\cite{zhang2021event} proposed the ANN-SNN hybrid network for event-based synthetic aperture imaging.

\begin{figure}
	\centering
	\includegraphics[width=\columnwidth]{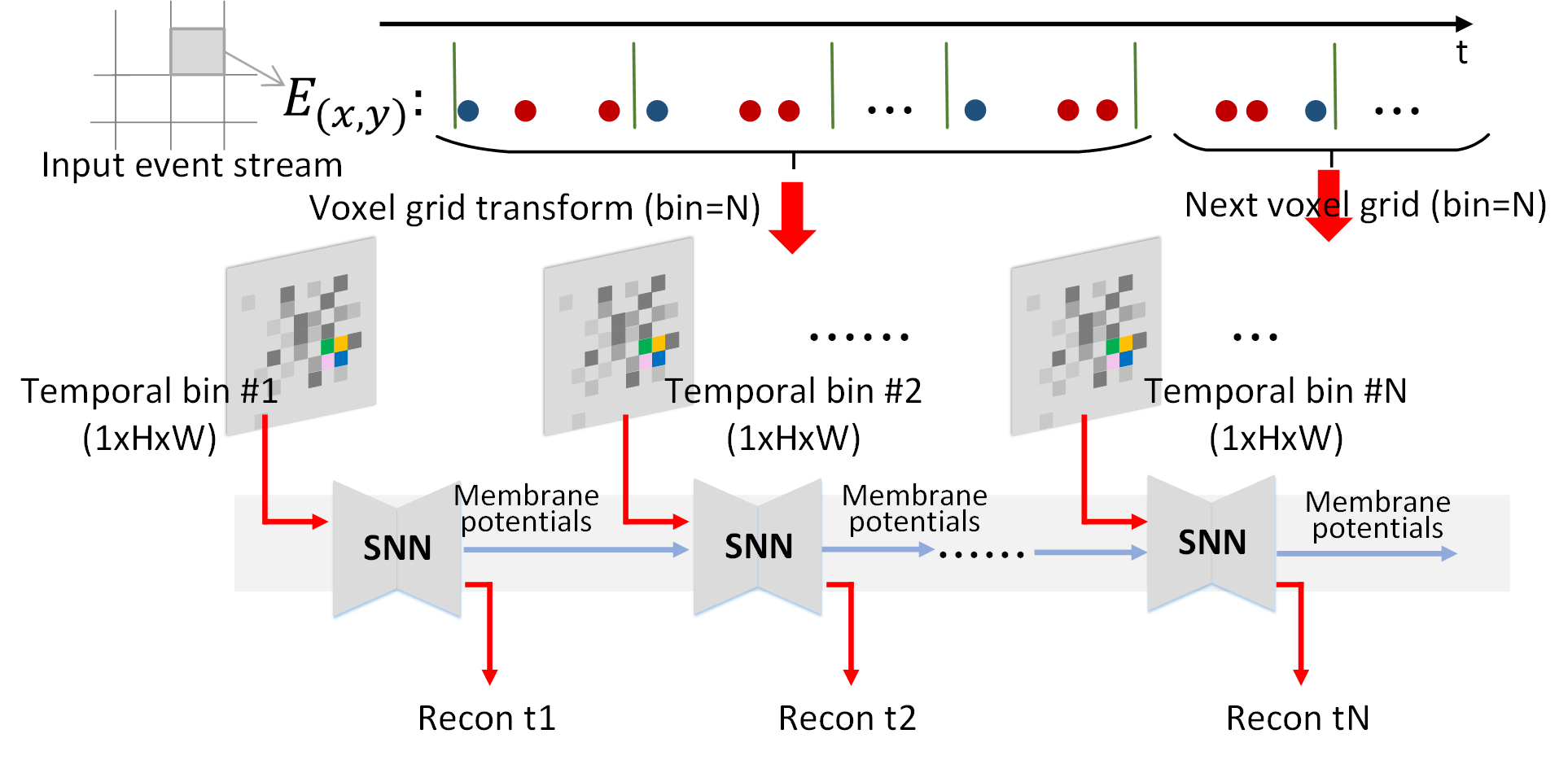} 
	\vspace{-5mm}
	\caption{\textbf{The event representation and work flow of our framework.} The event stream (red/blue dots represent on/off events, respectively) is split into multiple windows and transformed into continuous voxel grids.
	Each voxel grid includes $N$ temporal bins with different information. Our SNN recurrent uses the current single channel temporal bin and last membrane potentials of each spiking neuron to generate new reconstructions at each moment. }
	\label{fig1}
	\vspace{-4mm}
\end{figure}

\noindent\textbf{Event-based Video Reconstruction}
Video reconstruction is an important topic in event-based vision field.
Early reconstruction works are based on hand-crafted features to estimate intensity from events, e.g. optimization~\cite{bardow2016simultaneous}, regularization~\cite{munda2018real} and temporal filtering~\cite{scheerlinck2018continuous, scheerlinck2019asynchronous}.
Some works~\cite{cook2011interacting, kim2016real, rebecq2016evo} also applied SLAM to estimate the brightness.
Recently, deep learning methods have shown impressive performance on event-based video reconstruction.
Wang \emph{et al.}~\cite{wang2019event} utilized generative adversarial network (GAN) to reconstruct intensity with real grayscale frames.
Rebecq \emph{et al.}~\cite{rebecq2019events, rebecq2019high} proposed an effective E2VID model which based on a U-Net~\cite{ronneberger2015u} model. The network was trained in a supervised manner with a synthetic dataset generated from ESIM~\cite{rebecq2018esim}.
Scheerlinck \emph{et al.}~\cite{scheerlinck2020fast} proposed a light-weight framework to achieve fast inference speed with only a minor drop in accuracy.
Stoffregen \emph{et al.}~\cite{stoffregen2020reducing} proposed to use more complex synthetic dataset to train the network, bringing a large performance boost on real datasets.
Federico \emph{et al.}~\cite{paredes2021back} proposed a novel self-supervised learning method for image reconstruction, getting rid of training data.
Weng \emph{et al.}~\cite{weng2021event} presented a hybrid CNN-transformer network for image reconstruction.

In this work, different from the above ANN-based models, we first propose to use the energy-efficient deep SNN models for reconstructing videos from event stream.


\vspace{-1mm}

\section{Method}

\subsection{Input Representations}
To process the asynchronous event with SNN, the event data is needed to be converted into an event representation which includes the temporal information.
In this work, we use the continuous voxel grid~\cite{zhu2019unsupervised} to train and test our model, which is defined as: $E(x,y,t_n) = \sum_i p_i \max (0,1-|t_n-t_i^*|)$, where $t_i^* = \frac{B-1}{\Delta T}(t_i - t_0)$, $t_i^*$ is the normalized event timestamp. 
As shown in Fig.~\ref{fig1}, the event stream can be adaptively divided into continuous bins of voxel grid.

\subsection{Spiking Neurons}
\label{sec32}
ANN and SNN can model the same types of network topologies, but SNN replaces the artificial neuron model with a spiking neuron model. 
The artificial neuron model operates on a weighted sum of inputs, and passing the result through a sigmoid or ReLU nonlinearity. 
In SNN, the weighted sum of inputs contributes to the membrane potential of the spiking neuron.
If the membrane potential of the spiking neuron reaches a threshold, then the neuron will emit a spike to its subsequent connections. 
The information in SNN is propagated by discrete spikes,thus spiking neuron is the basic computing unit.

\noindent\textbf{LIF Neurons}
The Leaky Integrate-and-Fire (LIF) model~\cite{gerstner2014neuronal} is a widely used neuron model in SNN, which is more biologically realistic than the Integrate-and-Fire (IF) neuron model.
The subthreshold dynamics of LIF neuron is defined as $\tau \frac{{\rm d} V(t)}{{\rm d} t} = - (V(t) - V_{rest}) + X(t)$, 
where $V(t)$ represents the membrane potential of the neuron at time $t$, $X(t)$ represents the input to neuron, $\tau$ is the membrane time constant. A spike fires if $V(t)$ exceeds the threshold $V_{th}$, $V_{rest}$ is the resting potential after firing.
For better representation, we rewrite the above equation as the discrete form:
\vspace{-5mm}
\begin{eqnarray}
\vspace{-5mm}
\left\{
\begin{aligned}
&V_t = V_{t-1} + \frac{1}{\tau}(-(V_{t-1}-V_{rest})+X_t) \\
&S_t = H(V_t - V_{th})
\end{aligned}
\right.
\label{eq2}
\end{eqnarray}
where $V_t$ denotes the membrane potential after neuronal dynamics at $t$. $S_t$ denotes the spike output at $t$, $H(\cdot)$ denotes the Heaviside step function which is defined as $H(x) = 1$ for $x \geq 0$ and $H(x) = 0$ for $x < 0$. We set $V_{rest}$ = $V_{reset}$ in our work.
LIF neuron can extract temporal information during the integration and firing process, however, its output is binary spikes which can only represent limited information. Moreover, after each firing process, $M_t$ is reset thus the temporal information is also partially lost. Based on above, we introduce the membrane potential neurons.

\noindent\textbf{Membrane Potential Neurons}
The membrane potential neurons (MP neurons) are non-spiking neurons which output membrane potential instead of spikes~\cite{strohmer2021integrating, wu2021liaf}.
In our image reconstruction task, MP neurons can extract more useful temporal information hidden in the neurons.
The dynamics of the MP neurons is same as LIF neuron.
For MP neurons, $M_t$ is equal to $V_t$ since there is no spike fire and $V_{t}$ reset process. If we set $V_{rest}=0$, Eq.~\ref{eq2} can be written as:
\begin{eqnarray}
\vspace{-2mm}
\left\{
\begin{aligned}
&V_t =(1-\frac{1}{\tau}) V_{t-1} + \frac{1}{\tau}X_t \\
&O_t = V_t
\end{aligned}
\right.
\label{eq3}
\vspace{-5mm}
\end{eqnarray}
where $O_t$ denotes the output of the neuron at $t$. Eq.~\ref{eq3} is similar to the function of recurrent neural networks. The membrane time constant $\tau$ controls the balance between remembering $X_t$ and forgetting $V_{t-1}$. Thus it can be considered as a simple version of Long Short-Term Memory (LSTM) module~\cite{hochreiter1997long}.

\begin{figure}
	\centering
	
	\includegraphics[width=0.6\columnwidth]{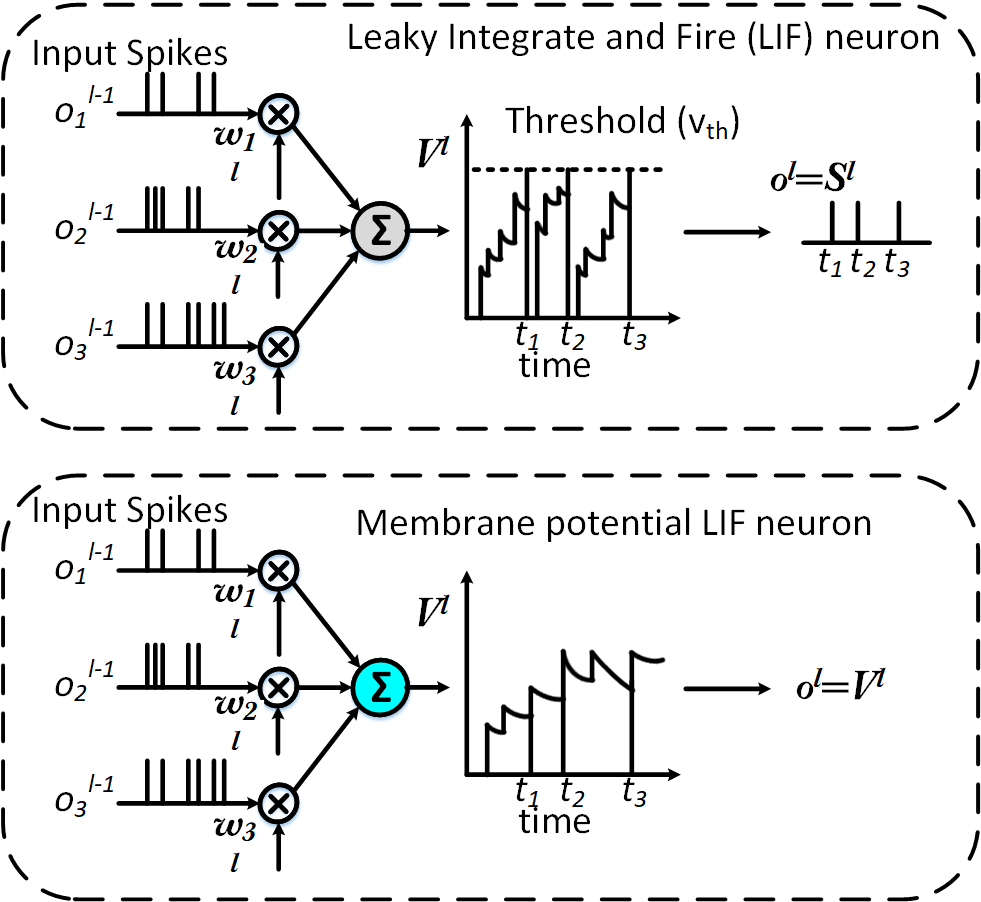} 
	\vspace{1mm}
	\caption{\textbf{The dynamics of LIF neuron and MP$\underline{~~}$LIF neuron.} For LIF neurons, if the membrane potential reaches a threshold, then the neuron will emit a spike to its subsequent connections and reset to resting state. At each time step, MP$\underline{~~}$LIF neuron outputs its membrane potential as the weighted sum of input spikes.}
	\vspace{-4mm}
	\label{fig2}
\end{figure}


\begin{figure*}
	\centering
	\includegraphics[width=1.9\columnwidth]{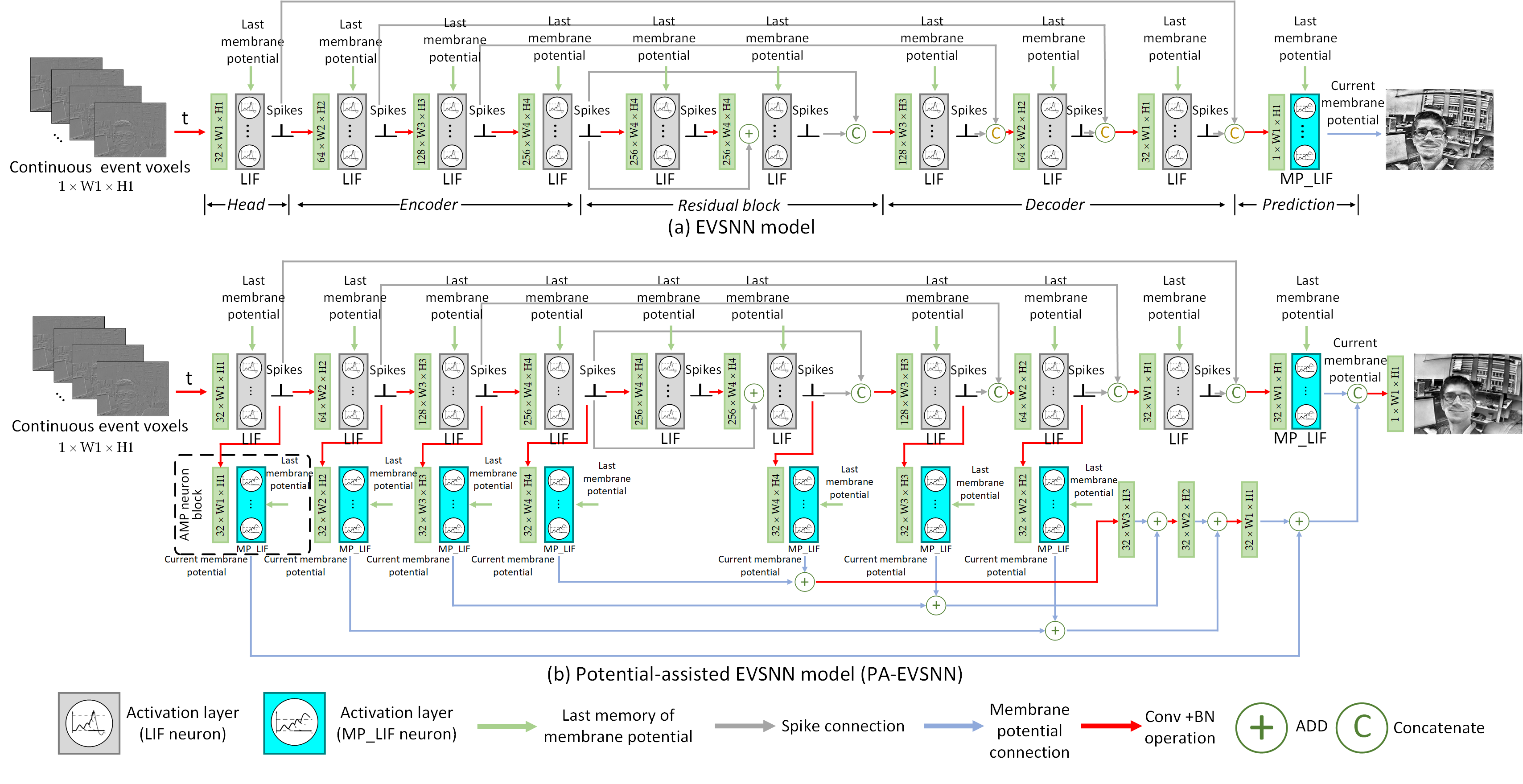} 
	\vspace{-1.5mm}
	\caption{\textbf{The proposed spiking neural network architecture.} EVSNN is a fully spiking neural network, composed of head, encoder, residual block, decoder, and prediction layers. Based on EVSNN as backbone, PA-EVSNN introduces MP neurons to further improve the performance. MP$\underline{~~}$LIF denotes MP neurons with the dynamics of LIF. As floating-point multiplication operations are introduced by MP neurons, we consider PA-EVSNN as a hybrid network. The energy consumptions of EVSNN and PA-EVSNN are 19.36 $\times$ and 7.75 $\times$ more computationally efficient than their ANN architectures. Please refer to our supplementary material for details of network architecture.}
	\vspace{-4mm}
	\label{fig3}
\end{figure*}

\subsection{The Proposed SNN Model}
In this paper, we propose two SNN architectures for event-based reconstruction, namely EVSNN and PA-EVSNN.
EVSNN is a fully spiking neural network, all synaptic operations in the network are SNN operations.
PA-EVSNN shares the same spiking encoder and decoder architecture, with the additional MP neurons to improve the performance.
Both models are fully convolutional networks, the architecture is shown in Fig.~\ref{fig3}.

\noindent\textbf{EVSNN (A fully spiking neural network)} Our EVSNN is a variant of the U-shaped model~\cite{ronneberger2015u}.
First, the event data is transformed into event voxels.
For each time step, a $1 \times W \times H$ event voxel is fed in to EVSNN and transformed as the size of $N_c \times W_1 \times H_1$, followed by $N_e$ encoder layers, $N_r$ residual blocks, $N_d$ decoder layers, and a final image prediction layer.
The number of channels is doubled after each encoder layer.
All spiking neurons in encoder layers, decoder layers, and residual blocks are LIF neurons, which enables computationally efficiency.
To ensure a fully SNN architecture, EVSNN utilizes concatenate as spike skip connection.
In the final image prediction layer, MP$\underline{~~}$LIF neuron is introduced to integrate all spikes and predict the gray scale image.
An ablation of each network component can be found in Sec.~\ref{sec44}. 
We use $N_c=32$, $N_e= N_d = 3$ and $N_r = 1$. 
EVSNN can handle most scenes in existing datasets while the computationally efficiency is 19.36 times than ANN architecture.

\noindent\textbf{PA-EVSNN (Potential-assisted EVSNN)}
EVSNN is a fully SNN with very low energy consumption.
However, the reconstruction performance is limited by the binary spikes (e.g., the gray scale of the image is not rich enough).
Based on EVSNN, we further propose a potential-assisted EVSNN model.
MP neuron is introduced in each encoder and decoder layer to help extract the temporal information hidden in the spikes.
We also propose an adaptive membrane potential (AMP) neuron, which greatly enhances the temporal receptive field of the network.
Notice that although the backbone of PA-EVSNN is SNN architecture, the introduction of MP neuron brings non binary spikes in the network (about 8.4$\%$ ANN floating-point operations), thus we consider PA-EVSNN as a hybrid network. Compared to existing ANN models, PA-EVSNN still has great advantages in energy consumption (7.75 $\times$ more efficient) while achieving comparable performance. More detailed analysis of the SNN and ANN operations can be found in Sec.~\ref{sec45}.

\noindent\textbf{Adaptive Membrane Potential Neurons}
As analyzed in Sec.~\ref{sec32}, the membrane time constant $\tau$ in MP neuron plays an analogous role as the gates in LSTM module. 
\cite{fang2021incorporating} proposed the parametric LIF neuron by introducing a learnable $\tau$ in classification task. 
Inspired by this, we propose an adaptive membrane potential neuron (AMP neuron).
Different from the fixed $\tau$ of parametric LIF neuron learned from the training dataset, AMP neuron can adjust $\tau$ according to the input spike to adapt to various reconstruction scenes.

\begin{figure}
	\centering
	\includegraphics[width=0.8\columnwidth]{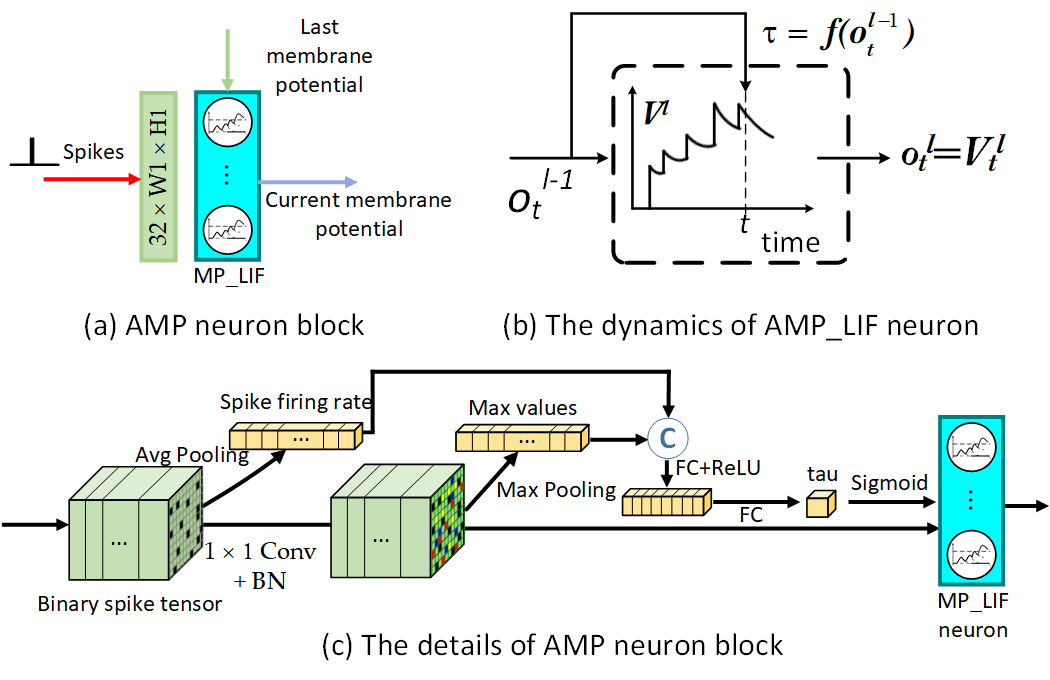} 
	\caption{\textbf{The adaptive membrane potential (AMP) neuron block.} (a) An AMP neuron block in Fig.~\ref{fig3}. (b) The dynamics of AMP$\underline{~~}$LIF neuron. The membrane time constant $\tau$ is adjusted by the input $o_t^{l-1}$. (c) The details of AMP neuron block.}
	\vspace{-5mm}
	\label{fig4}
\end{figure}

According to Eq.~\ref{eq3}, ideally, when the light changes fast, the network should choose a large $\tau$ to remember more new information while forgetting more last memory, and vice versa.
Due to the event measures the change of light intensity, the average spike firing rate reflects the global motion of the scene to a certain extent, which is useful for estimating a proper $\tau$.
The spike firing rate of each channel in $l$-th layer can be estimated by ${\rm F} = {\rm AvgPool}(S_{l})$, where ${\rm AvgPool}(\cdot)$ denotes the average pooling operation, $S_l$ is the spike tensor of $l$-th layer. Then the local motion intensity of input spikes can be estimated by ${\rm I} = {\rm MaxPool}({\rm Conv}(S_l))$, where ${\rm MaxPool}(\cdot)$ denotes the max pooling operation.
Finally, the membrane time constant is updated by
\begin{eqnarray}
\vspace{-2mm}
\tau = \frac{1}{\mathcal{S}({\rm Linear}([{\rm F},{\rm I}]))}
\label{eq4}
\vspace{-6mm}
\end{eqnarray}
where $\mathcal{S}(\cdot)$ denotes the sigmoid activation function, ${\rm Linear}(\cdot)$ is the full connection layer shown in Fig.~\ref{fig4}.

\noindent\textbf{Loss Functions}
We use LPIPS loss and temporal consistency loss:
$\mathcal{L}_{total} = \sum_{k=0}^L \mathcal{L}_{k}^R + \lambda  \sum_{k=L_0}^L \mathcal{L}_{k}^{TC}$, 
where $\mathcal{L}_{k}^R$ is the LPIPS loss~\cite{zhang2018unreasonable}, $\mathcal{L}_{k}^{TC}$ is temporal consistency loss~\cite{lai2018learning, rebecq2019high}.

\subsection{Training Details of SNN}
During the training process, we set $L$ in loss function same as the training sequence length (i.e., 40 - 60), and $L_0$ is set as 2.
In each time step, an event voxel with the size of $1\times H_1 \times W_1$ is fed into the network.
According to Eq.~\ref{eq5}, the backpropagated errors pass through the spiking neuron layer and MP neuron layer using BackPropagation Through Time (BPTT)~\cite{werbos1990backpropagation}. 
In BPTT, the network is unrolled for all discrete time-steps. 
The loss is calculated every 5 time steps, and the weight update is computed as the sum of gradients from each time-step as follows:
\vspace{-2mm}
\begin{eqnarray}
&&\label{eq5}\Delta w ^l = \sum_n \frac{\partial \mathcal{L}_{total}}{\partial o^l_t}  \frac{\partial o^l_t}{\partial V^l_t} \frac{\partial{V}^l_t}{\partial w^l}  \\
&&{\rm where} \ \  \frac{\partial o^l_t}{\partial V^l_t} = \left\{
\begin{aligned}
&H_1^{'}(V_t-V_{th}) \ \   {\rm if} \ \ o^l_t=S^l_t \\
&1  \ \ \ \ \ \ \ \ \ \ \ \ \ \ \ \ \ \ \ \ \ \ \   {\rm if} \ \ o^l_t = V^l_t
\end{aligned}
\right.\nonumber
\end{eqnarray}

\vspace{-2mm}
\noindent where $o^l_t$ is the output of the neuron at time $t$, $\frac{\partial o^l_t}{\partial V^l_t}$ denotes the derivative of spike with respect to the membrane potential after charging at time step $t$. Since $\frac{\partial o^l_t}{\partial V^l_t}$ is not differentiable, we adopt surrogate gradient method~\cite{neftci2019surrogate} to calculate it.
The shifted ArcTan function $H_1(x) = \frac{1}{\pi}\arctan(\pi x) + \frac{1}{2}$ is utilized as the surrogate function of the Heaviside step function $H(\cdot)$. If the neuron is a spiking neuron, we have . Otherwise, if the neuron is a MP neuron with no spiking output, then $o^l_t = V^l_t$, we have $\frac{\partial o^l_t}{\partial V^l_t} = 1$ which is similar to an ANN activation function.

\begin{figure}
	\centering
	\includegraphics[width=\columnwidth]{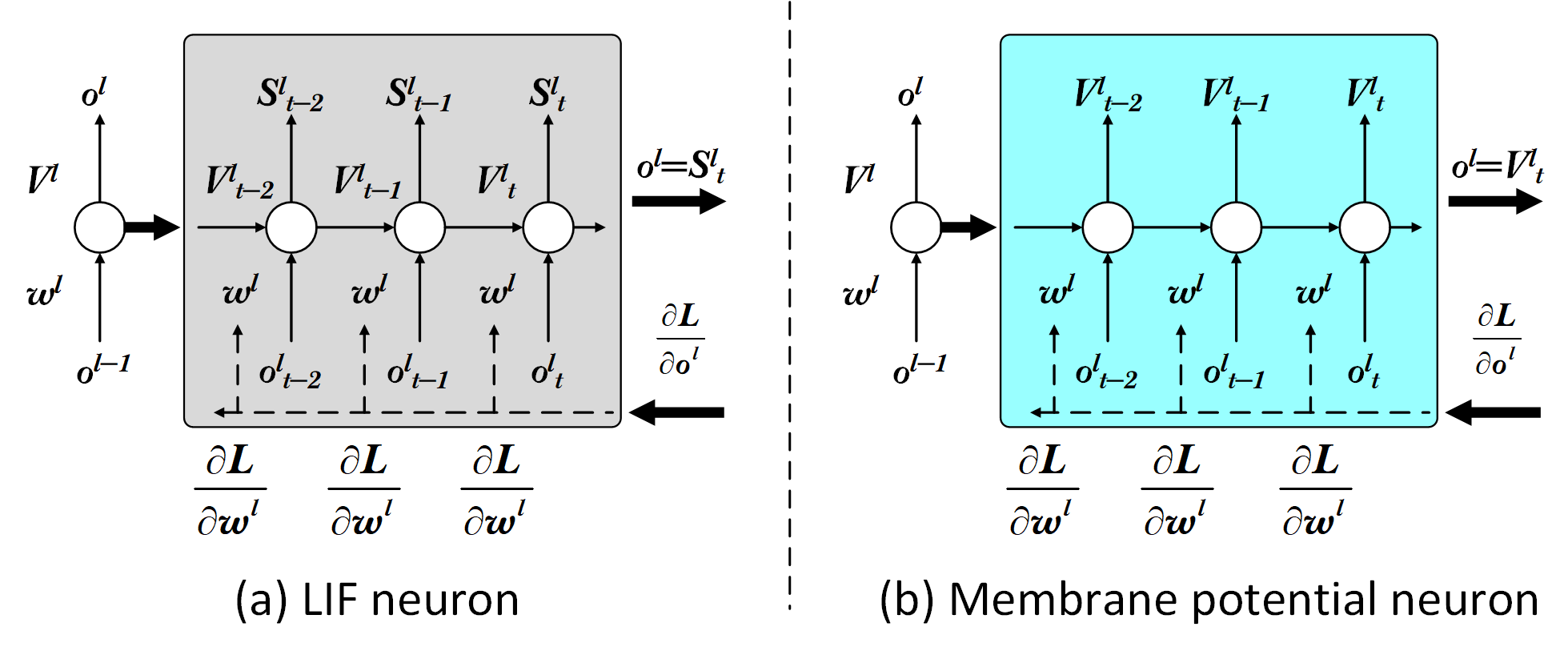} 
	\vspace{-5mm}
	\caption{\textbf{The backpropagation of spiking neurons.} For LIF neurons, we use ArcTan as surrogate function to calculate the derivative of spiking function. For MP neurons, the gradients can be directly computed by Eq.~\ref{eq5}. }
	\label{fig5}
	\vspace{-5mm}
\end{figure}

\begin{figure*}
	\centering
	\includegraphics[width=1.4\columnwidth]{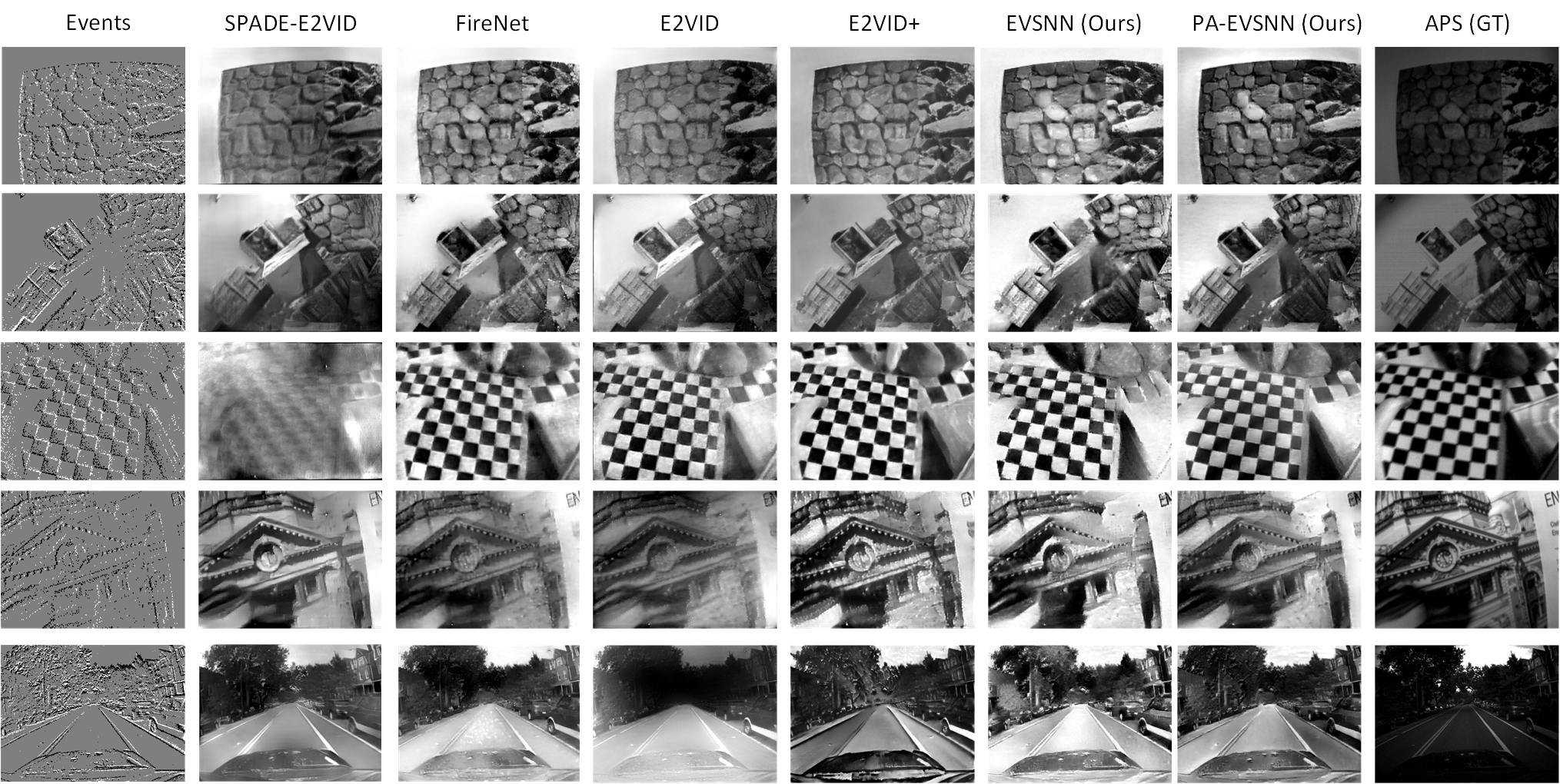} 
	\caption{\textbf{Qualitative comparison with state-of-the-art ANN-based methods.} We compare our SNN models with four ANN-based models (SPADE-E2VID, FireNet, E2VID, and E2VID+) on IJRR (Row 1-2), HQF (Row 3-4), and MVSEC (Row 5) datasets. The results show that the proposed EVSNN and PA-EVSNN perform comparably to most ANN-based models, and the energy consumptions are 24.15 times and 8.76 times lower than E2VID, respectively (see Table~\ref{table7}). More qualitative results can be found in our supplementary material. }
	\label{fig6}
\end{figure*}

\begin{table*}
	\scriptsize
	\caption{Comparison on IJRR, HQF, and MVSEC Datasets.}
	\begin{center}
		\begin{tabular}{lccccccccc}
			\toprule[1pt]
			\multirow{2}{*}{Method}&  \multicolumn{3}{c}{IJRR}& \multicolumn{3}{c}{MVSEC} & \multicolumn{3}{c}{HQF} \\
			\cmidrule(r){2-4} \cmidrule(r){5-7}\cmidrule(r){8-10}
			& MSE $\downarrow$&SSIM $\uparrow$&LPIPS $\downarrow$& MSE $\downarrow$&SSIM $\uparrow$&LPIPS $\downarrow$& MSE $\downarrow$&SSIM $\uparrow$&LPIPS $\downarrow$ \\  \hline
			$^*$E2VID&0.059&\textbf{0.643}&\underline{0.338}&0.138&0.377&0.651&0.081&\underline{0.545}&0.406\\
			$^*$FireNet&0.060&0.602&0.340&0.105&0.361&0.600&0.065&0.542&\underline{0.391}\\
			$^*$SPADE-E2VID&0.063&0.572&0.365&\underline{0.095}&\textbf{0.443}&0.556&0.080&0.512&0.424\\ 
			$^*$$^1$E2VID+&\textbf{0.043}&0.618&\textbf{0.321}&\textbf{0.088}&\underline{0.427}&\textbf{0.490}&\textbf{0.047}&\textbf{0.560}&\textbf{0.338}\\ \hline
			$^\dag$EVSNN (Ours)&0.061&0.570&0.362&0.104&0.389&\underline{0.538}&0.086&0.482&0.433\\
			$^\dag$PA-EVSNN (Ours)&\underline{0.046}&\underline{0.626}&0.367&0.107&0.403&0.566&\underline{0.061}&0.532&0.416\\  
			
			\toprule[1pt]
		\end{tabular}
		\begin{tablenotes}
			\scriptsize
			\item[*]$\ \ \ \ \ \ \ \ \ \ \ \ \ \ \  $$^1$ E2VID+ is trained on the simulated dataset proposed in~\cite{stoffregen2020reducing} , while other five models are all trained on the simulated dataset from~\cite{rebecq2019high}.\\
			$\ \ \ \ \ \ \ \ \ \ \ \ \ \ \   $$^*$ ANN model.   $^\dag$ SNN model. Notice that the energy consumption of SNN is much lower than that of ANN, see Table~\ref{table7} for detail.
		\end{tablenotes}
	\end{center}
	\vspace{-8mm}
	\label{table1}
\end{table*}

\vspace{-2mm}
\section{Experiments}
\subsection{Experimental setup}
For fair comparison to ANN-based reconstruction methods, we use the exact same synthetic data from E2VID~\cite{rebecq2019high} to train our SNN.
The dataset is generated by ESIM, an event simulator, and consists of 950 training sequences and 50 validation sequences.
MS-COCO images~\cite{lin2014microsoft} are mapped to a 3D plane and random 6-DOF camera motions are used to trigger events.
During training, the nonzero values of event tensors are normalized as mean and standard deviation is 0 and 1, respectively.
The events and images are randomly cropped to 128$\times$128 to augment the data.

Our models are implemented with SpikingJelly~\cite{fang2020other}, an open-source deep learning framework for SNNs based on PyTorch~\cite{paszke2019pytorch}. 
An NVIDIA TITAN Xp GPU is used to train our model.
We adopt the batch size of 8 and Adam optimizer~\cite{kingma2014adam} in the training process. 
The network is trained for 100 epochs, with a learning rate of 0.002. 
The weight $\lambda$ of temporal consistency loss is set as 1.
The reset value $V_{reset}$ of all neurons is set to 0, and the membrane time constant $\tau$ of LIF neurons is set to 2.

\subsection{Evaluation on Public Datasets}
We evaluate our model on three public datasets IJRR~\cite{mueggler2017event}, MVSEC~\cite{zhu2018multivehicle}, and HQF~\cite{stoffregen2020reducing}.
Following~\cite{rebecq2019high} and~\cite{scheerlinck2020fast}, to ensure the intensity values lay within a similar range, we apply histogram normalization to both the output and groundtruth frames.
Moreover, to make the timestamps of the reconstruction and groundtruth strictly consistent, we use the events between two adjacent frames to generate each reconstruction.
We compare our models with four state-of-the-art-methods E2VID~\cite{rebecq2019high}, FireNet~\cite{scheerlinck2020fast}, SPADE-E2VID~\cite{cadena2021spade}, and E2VID+~\cite{stoffregen2020reducing}.
All results are generated by the pre-trained model from the original paper.
We compared reconstructed images against groundtruths using the metrics: mean squared error (MSE), structural similarity (SSIM)~\cite{wang2004image} and perceptual similarity (LPIPS)~\cite{zhang2018unreasonable}.

The main quantitative results are presented in Table~\ref{table1}. Notice that E2VID+ is trained on a more challenging synthetic dataset, while the other five methods are trained on the same data as~\cite{rebecq2019high}.
To sum up, E2VID+ performs best in most datasets.
SPADE-E2VID performs well on MVSEC dataset, but the SSIM and LPIPS are lower than E2VID on IJRR and HQF datasets.
The results show that EVSNN can handle these scenes.
Our PA-EVSNN achieves comparable performance to ANN-based models such as E2VID and FireNet.
Please refer to our supplementary material for additional quantitative and qualitative results.


\subsection{Temporal Component Ablation}
\begin{figure*}
	\centering
	\includegraphics[width=1.6\columnwidth]{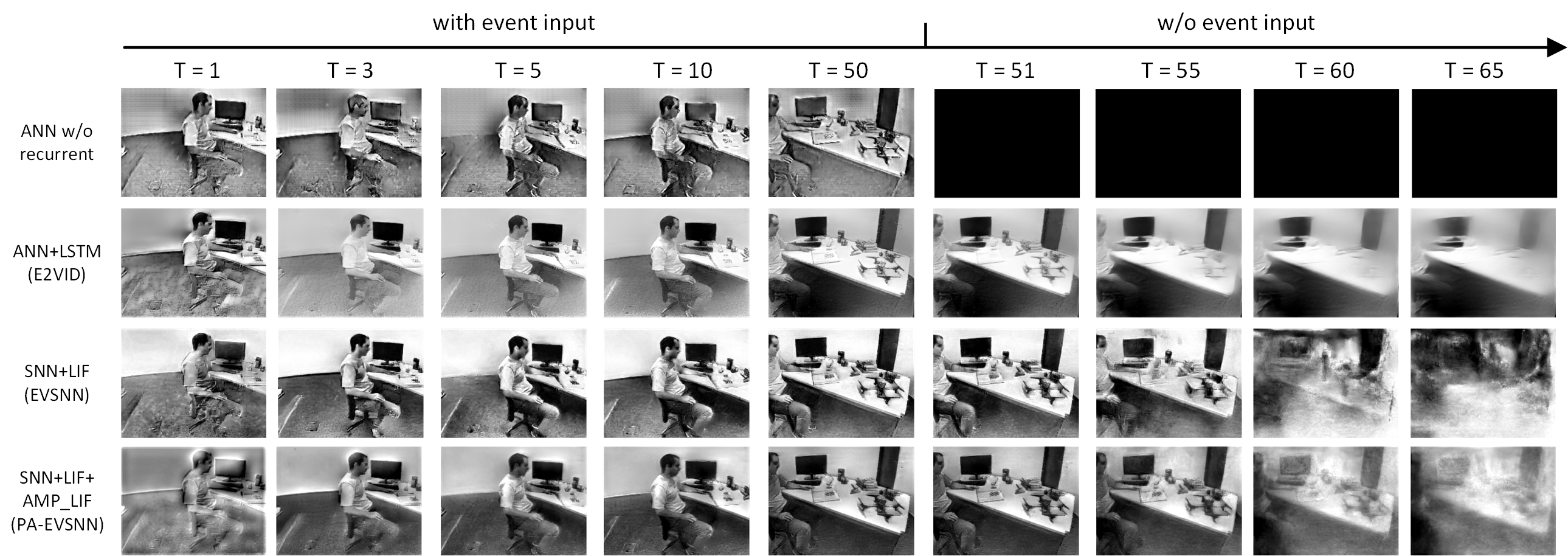} 
	\caption{\textbf{Comparison on the different temporal components of SNN and ANN.} 
	This figure shows image reconstructions of different ANN and SNN variants at initialization and end phases.
	At initialization phase (T = 1-50), the states of temporal components (e.g., LSTM and spiking neuron) are initialized at zero, all models are fed continuous event tensors to test the reconstruction of each moment.
	In subsequent iterations after T = 50, the models are fed empty event tensors to test the effective size of temporal receptive field.}
	\label{fig7}
\end{figure*}

Inspired by~\cite{rebecq2019high}, we design an experiment to measure the effective size of the temporal receptive field of SNN and ANN.
As shown in Fig.\ref{fig7}, four different settings are tested: ANN w/o recurrent, ANN + LSTM (E2VID), SNN + LIF (EVSNN), and SNN + LIF + AMP$\underline{~~}$LIF.
These networks are all based on U-Net architecture with three encoders.
To verify the ability of spiking neurons in temporal information extraction, at initialization phase (T = 1-50), the states of temporal components (e.g., LSTM and spiking neuron) are initialized at zero.
The images at each moment are reconstructed by continuous event input.
Then we artificially stop the events at T=50.
In subsequent iterations after T = 50, we feed the empty event tensors to the network and reconstruct images to test the effective size of temporal receptive field.

\begin{figure*}
	\centering
	\includegraphics[width=2\columnwidth]{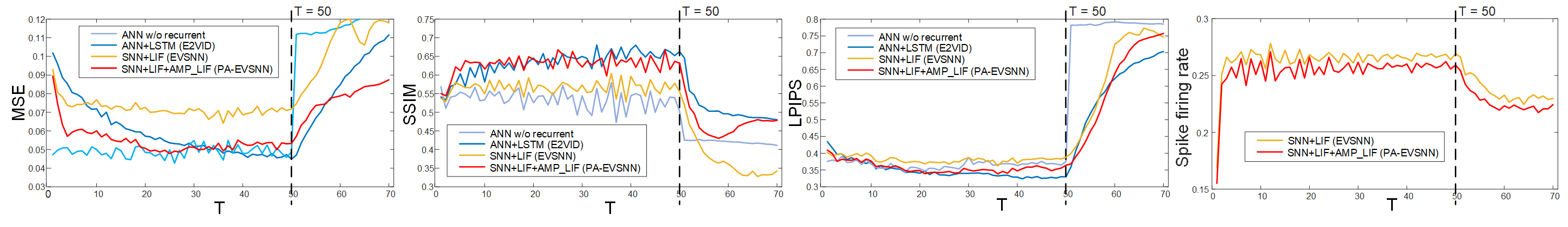} 
	\vspace{-2mm}
	\caption{\textbf{Quantitative analysis of the temporal components.} This figure shows the MSE, SSIM, LPIPS, and spike firing rate at each iteration. The experiment setting is same as Fig.\ref{fig7}, four ANN and SNN variants (ANN w/o recurrent, ANN + LSTM, SNN + LIF, and SNN + LIF + AMP$\underline{~~}$LIF) are tested. The results show that spiking neuron can improve the ability of temporal information extracting.}
	\vspace{-3mm}
	\label{fig8}
\end{figure*}

To better analyze the results, we randomly pick 50 event sequences from dynamic$\underline{~~}$6dof of IJRR dataset, and plot the average values of MSE, SSIM, LPIPS in Fig.~\ref{fig8}.
The spike firing rates of EVSNN and PA-EVSNN are also reported.
The results shown in Fig~\ref{fig7} and~\ref{fig8} show that E2VID, EVSNN, and PA-EVSNN can complete initialization in 10 iterations. As T increases, the quantitative results of E2VID, EVSNN, and PA-EVSNN continue to improve, which shows the effectiveness of temporal component.
In contrast, the quantitative scores of ANN w/o recurrent do not change significantly, since it has no temporal component.
After T=50, the quantitative score of E2VID, EVSNN, and PA-EVSNN decreases slowly, indicating that our SNN model has temporal receptive field similar to ANN + LSTM.
Assisted by the membrane potential, PA-EVSNN performs better than EVSNN on quantitative scores.
To sum up, our SNN structure has the capacity of temporal information extracting although it may be weaker than ANN+LSTM.



\subsection{Spiking Neural Network Architecture}
\label{sec44}
We investigate different SNN network architectures.
All the experiments are conducted on IJRR dataset.

\noindent\textbf{Spiking Neurons} 
In the first ablation study, we explore the effect of different spiking neurons on the reconstruction performance.
We test three types of spiking neurons in EVSNN: IF neuron, LIF neuron, and PLIF (parametric LIF) neuron.
Since IF neuron simply integrates inputs and lacks the decay mechanism, its performance is worse than the other two neurons.
For the other two spiking neurons, a membrane time constant controls the decay. 
As shown in the upper part of Table~\ref{table2}, LIF neurons perform slightly better than PLIF neurons. 

\noindent\textbf{Membrane Potential Neurons} Based on the EVSNN-LIF architecture, we further analyze the effects of MP neurons.
Fig.~\ref{fig7} and~\ref{fig8} show that MP neurons can improve the reconstruction quality of SNN. 
We test four types of MP neurons: MP$\underline{~~}$IF, MP$\underline{~~}$LIF, MP$\underline{~~}$PLIF, and AMP$\underline{~~}$LIF.
These neurons are non-spiking neurons which output membrane potential instead of spikes.
PLIF neurons can learn a fixed membrane time constant based on the training dataset. 
However, the fixed membrane time constant learned from synthetic data may be not suitable for complex scenes.
As shown in the lower part of Table~\ref{table2}, our AMP$\underline{~~}$LIF performs best because it can adaptively adjust the decay rate by the input spikes.

\noindent\textbf{Spike Skip Connection} Spike skip connection gathers the spike outputs of the encoder and decoder.
An effective connection operation can greatly improve the performance of SNN.
Based on the EVSNN-LIF and PA-EVSNN-AMP$\underline{~~}$LIF architectures, we study four types of spike connections.
As shown in Table~\ref{table3}, ADD performs best because it retains more information by adding spikes of the encoding and decoding layers. However, it brings non-spike output, e.g., the addition of two spikes will output 2, which breaks the fully SNN architecture and adds additional power consumption.
In contrast, OR, IAND, and CONCAT all output spikes. As shown in Table~\ref{table3}, CONCAT performs best while the IAND also performs well. 
Although the number of parameters of CONCAT is twice that of IAND, we choose CONCAT in our architecture for better performance.

\begin{table}
	\scriptsize
	\caption{Ablation studies of spiking neurons and MP neurons.}
	\vspace{-4mm}
	\begin{center}
		\begin{tabular}{lccc}
			\toprule[1pt]
			Model& MSE$\downarrow$ &SSIM$\uparrow$ & LPIPS$\downarrow$\\
			\hline
			EVSNN-IF  & 0.108&0.341&0.608  \\
			EVSNN-PLIF  & 0.063&0.569&0.367  \\
			EVSNN-LIF  & \textbf{0.061}&\textbf{0.570}&\textbf{0.362}  \\ \hline
			PA-EVSNN-MP$\underline{~~}$IF  &0.121 &0.362 & 0.741\\
			PA-EVSNN-MP$\underline{~~}$LIF  &0.056 &0.597  & 0.388\\
			PA-EVSNN-MP$\underline{~~}$PLIF  &0.053 &0.599  & 0.378\\
			PA-EVSNN-AMP$\underline{~~}$LIF  &\textbf{0.042}&\textbf{0.632}  & \textbf{0.376}\\
			\toprule[1pt]
		\end{tabular}
	\end{center}
	\vspace{-7mm}
	\label{table2}
\end{table}
\begin{table}
	\scriptsize
	\caption{Ablation studies of different spike skip connections.}
	\vspace{-3mm}
	\begin{center}
		\begin{tabular}{lccc}
			\toprule[1pt]
			Model& MSE$\downarrow$ &SSIM$\uparrow$ & LPIPS$\downarrow$\\
			\hline
			$^1$EVSNN-ADD  & \textbf{0.049}&\textbf{0.586}&\textbf{0.350}  \\
			$^2$EVSNN-OR  & 0.063&0.534&0.395  \\ 
			$^3$EVSNN-IAND  & \underline{0.051}&0.557&\underline{0.357} \\ 
			$^4$EVSNN-CONCAT  &0.061 &\underline{0.570}&0.362  \\\hline
			$^1$PA-EVSNN-ADD  &\textbf{0.041} &\textbf{0.635}  & \underline{0.388}\\
			$^2$PA-EVSNN-OR  &0.064 &0.591  & 0.436\\
			$^3$PA-EVSNN-IAND  &0.055 &0.602  & 0.410\\
			$^4$PA-EVSNN-CONCAT  &\underline{0.046} &\underline{0.626}  & \textbf{0.367}\\ 
			\toprule[1pt]
		\end{tabular}
		\begin{tablenotes}
			\scriptsize
			\item[*] Defining the connect operation as $g(A_l, B_l)$, where $A_l$ $\in$ $\{$0,1$\}$ and $B_l$ $\in$ $\{$0,1$\}$ denote the spike output of $l$-th encoder and decoder, respectively. The different connections can be implemented as
			\item[*]$^1$ ADD: $g_{ADD}(A_l, B_l) = A_l + B_l$\\
			\item[*]$^2$ OR: $g_{OR}(A_l, B_l) = \max (A_l, B_l)$\\
			\item[*]$^3$ IAND: $g_{IAND}(A_l, B_l) = (1 - A_l) \cdot B_l$\\
			\item[*]$^4$ CONCAT: $g_{CON.}(A_l, B_l) = [A_l, B_l]$\\
		\end{tablenotes}
	\end{center}
	\vspace{-10mm}
	\label{table3}
\end{table}

\noindent\textbf{Number of Encoders and Residual blocks} Finally, we search the number of encoders and residual blocks. 
The results are shown in Table~\ref{table4}, e.g., EVSNN-e3-res1 means EVSNN with three encoders and one residual block. Considering both the performance and complexity, we choose EVSNN-e3-res1 and PA-EVSNN-e3-res1 as our models.


%
\vspace{0mm}
\subsection{Energy Consumption and Limitation Analysis}
\label{sec45}

\noindent\textbf{Energy Comparison of SNN and ANN}
Typically, the number of synaptic operations is used as a metric for benchmarking the computational energy of neuromorphic hardware~\cite{merolla2014million}.
In ANN, each operation computes a dot-product involving one floating-point (FP) multiplication and one FP addition as a multiply-accumulate (MAC) computation. 
In contrast, the computations in SNN implemented on neuromorphic hardware are event-driven. Therefore, in the absence of spikes, there are no computations and no active energy is consumed~\cite{davies2018loihi}.
Thus, in SNN, each operation is only one FP addition due to binary spikes. 
The low consumption of SNN synapse operation combined with activation sparsity provides large improvements in computational efficiency.
\begin{table}
	\scriptsize
	\caption{Ablation studies of different network architectures.}
	\vspace{-3mm}
	\begin{center}
		\begin{tabular}{lccc}
			\toprule[1pt]
			Model& MSE$\downarrow$ &SSIM$\uparrow$ & LPIPS$\downarrow$\\
			\hline
			EVSNN-e2-res1 & \textbf{0.060}&0.569&0.364  \\
			EVSNN-e3-res1  & \underline{0.061}&\underline{0.570}&\underline{0.362} \\
			EVSNN-e4-res1  & \underline{0.061}&\textbf{0.576}&0.379  \\ \hdashline
			EVSNN-e3-res0  &\underline{0.061}&0.569&\textbf{0.360}  \\
			EVSNN-e3-res2  & 0.067&\underline{0.570}&0.371  \\ \hline
			PA-EVSNN-e2-res1  &0.050 &\textbf{0.628}  & 0.376\\
			PA-EVSNN-e3-res1  &\textbf{0.046} &\underline{0.626}  & \underline{0.367}\\
			PA-EVSNN-e4-res1  &0.048 &0.618  & 0.379\\\hdashline
			PA-EVSNN-e3-res0  &0.058 &0.599  & 0.413\\
			PA-EVSNN-e3-res2  &\textbf{0.046} &0.615  & \textbf{0.361}\\
			\toprule[1pt]
		\end{tabular}
	\end{center}
	\vspace{-4mm}
	\label{table4}
\end{table}

\begin{table}
	\scriptsize
	\caption{Spike firing rate of EVSNN and PA-EVSNN.}
	\begin{center}
		\begin{tabular}{lcccc}
			\toprule[1pt]
			\multirow{2}{*}{Layer}& \multirow{2}{*}{\shortstack{Spiking Neuron\\ Num.}}&\multirow{2}{*}{\shortstack{Neuron\\ Type}} & \multicolumn{2}{c}{Spike Firing Rate} \\
			\cmidrule(r){4-5}
			& & &EVSNN&PA-EVSNN \\ \hline
			Head  &32$\times$H$\times$W& LIF &0.2479&0.2444 \\ 
			Down1  &64$\times$H$\times$W& LIF  &0.2459  & 0.2308\\
			Down2  &128$\times$H$\times$W &LIF  &0.1352  & 0.1339\\
			Down3  &256$\times$H$\times$W &LIF &0.1174 & 0.1183\\
			Res1-1  &256$\times$H$\times$W& LIF  &0.1241 & 0.1098\\
			Res1-2  &256$\times$H$\times$W& LIF  &0.1308  & 0.1200\\
			Up1  &128$\times$H$\times$W& LIF  &0.1905  & 0.1983\\
			Up2  &64$\times$H$\times$W &LIF  &0.3338  & 0.3573\\
			Up3  &32$\times$H$\times$W& LIF  &0.3580  & 0.3081\\\hline
			\multicolumn{3}{c}{Overall spike firing rate}&0.2642&0.2511\\
			\toprule[1pt]
		\end{tabular}
	\end{center}
	\vspace{-7mm}
	\label{table5}
\end{table}

To compare the consumption between SNN and ANN architectures, the evaluation should be conducted on the same structure~\cite{rathi2021diet}.
Thus, we compute the energy consumption between our SNN models and their ANN versions (e.g., replace the spiking neurons with ReLU). 
In most technologies, the addition operation is much cheaper than the multiplication operation.
We compute the energy cost/operation for ANNs and SNNs in 45nm CMOS technology. The energy cost for 32-bit ANN MAC operation is 5.1 more than SNN addition operation (4.6pJ vs. 0.9pJ)~\cite{horowitz20141}.

The number of synaptic operations in SNN can be calculated by multiplying $\#OP_{ANN}$\footnote{In ANN-based model, the number of ANN operations (MAC) is defined by $\# OP_{ANN} = \sum k_w \times k_h \times c_{in} \times h_{out} \times w_{out} \times c_{out}$, where $k_w$ and $k_h$ are kernel size, $c_{in}$ and $c_{out}$ are the number of input and channels, $h_{out}$ and $w_{out}$ are output feature map size, and $f_{in}$ and $f_{out}$ is the number of input (output) features.} by the spike firing rate.
For example, a spike rate of 1 (every neuron fired) implies that the number of operations for ANN and SNN are the same (though operations are MAC in ANN while addition in SNNs). 
Lower spike rates denote more sparsity in spike events and higher energy-efficiency.
As shown in Table~\ref{table5}, we count the average spike firing rate of EVSNN and PA-EVSNN on IJRR dataset. 
The comparison results are shown in Table~\ref{table6}.
Notice that our models do not require multiple time-steps simulation, which brings a great advantage in energy consumption.
For EVSNN, all operations are SNN operations, the average spike firing rate is 26.4$\%$, its energy consumption is 19.36 $\times$ lower than ANN.
Since there are 8.4$\%$ MAC operations in PA-EVSNN, the average spike firing rate of spiking neurons is 25.1$\%$, it costs 7.75 $\times$ lower energy consumption compared to its fully ANN version.

\noindent\textbf{Energy Comparison with E2VID}
Here we compare the energy consumption of our models with E2VID.
Table~\ref{table7} reports the energy comparsion\footnote{Energy = $\#OP_{ANN}\times 4.6 {\rm pJ} + \#OP_{SNN} \times {\rm 0.9pJ} \times { SpikeRate}$. Notice that $\#OP_{SNN}$ must be operated on binary spikes (i.e., 0 or 1).} with 180$\times$240 input size.
Each ANN operation consumes 4.6 pJ, brings 20.07G$\times$4.6pJ = 9.232$\times$10$^{-2}$J energy consumption.
Compared to LSTM, GRU is a recurrent module with fewer parameters.
Our EVSNN has 16.12G SNN operation with 26.4$\%$ spike firing rate, which costs 16.12G$\times$26.4$\%$$\times$0.9pJ = 3.83$\times 10^{-3}$J. 
For PA-EVSNN, we consider the 1.49G operations which come from MP neurons as ANN operations.
Thus, the overall energy cost of PA-EVSNN is 1.49G$\times$4.6pJ+16.35G$\times$25.1$\%$$\times$0.9pJ = 1.055$\times$10$^{-2}$J.
In summary, the energy consumptions of EVSNN and PA-EVSNN are 24.15 $\times$ and 8.76 $\times$ more computationally efficient than E2VID, respectively.

\noindent\textbf{Limitation}
To make SNN training faster and more stable, we add the batch normalization (BN) after convolution (CONV) layer.
Notice that BN can be folded in a CONV layer after training~\cite{rueckauer2017conversion}. 
However, BN is not unbiased. If there is no spike input, BN will also produces non-zero values, which may activate spiking neurons. This will increase the spike rate of SNN, thereby increasing energy consumption (see Fig.~\ref{fig8}, the spike rate $>$ 0 when no event input (T $>$ 50)). Reducing the spike rate may be a future direction.

%
%

\begin{table}
	\scriptsize
	\caption{Comparison of compute energy between ANN and SNN.}
	\vspace{-3mm}
	\begin{center}
		\begin{tabular}{lcccc}
			\toprule[1pt]
			 &  EVSNN&PA-EVSNN\\ 
			\hline
			$^1${\makecell[l]{(a) Normalized $\# OP_{ANN}$}} & 1&1\\ 
			$^2${\makecell[l]{(b) Normalized $\# OP_{SNN}$}}  &0.264 &0.251  \\
			$^3$(c) Normalized $\# OP_{MP\_{layer}}$  &0 &0.084  \\
			$^4$(d) {\makecell[l]{ANN/SNN Energy }}& 19.36 &7.75 \\
			\toprule[1pt]
		\end{tabular}
		\begin{tablenotes}
	\scriptsize
	\item[*]$^1$ $\# OP_{ANN}$ is the total number of ANN operations if all spiking neurons are replaced with an ANN activation function (e.g., ReLU).\\
	$^2$ $\# OP_{SNN} = SpikeRate \times \#OP_{ANN}$.   \\
	$^3$ MP layer contains FP multiplications and additions, thus its consumption is considered same as ANN. \\
	$^4$ Each operation in ANN (SNN) consumes 4.6pJ (0.9pJ). ANN/SNN Energy can be calculated by $\frac{(a)\times 4.6}{(c) \times 4.6+(1-(c)) \times (b) \times 0.9}$.
\end{tablenotes}
	\end{center}
	\vspace{-5mm}
	\label{table6}
\end{table}

\begin{table}
	\scriptsize
	\caption{Energy comparison of E2VID and our models.}
	\begin{center}
		\begin{tabular}{lcccc}
			\toprule[1pt]
			& \multirow{2}{*}{\shortstack{E2VID\\-LSTM}} & \multirow{2}{*}{\shortstack{E2VID\\-GRU}} & \multirow{2}{*}{EVSNN} & \multirow{2}{*}{PA-EVSNN}\\
			&&&&\\ \hline
			{Para. Num.}& 10.71M & 9.16 M  & 4.41M   & 4.62M  \\
			{Spike Rate} & - & -&0.264 & 0.251 \\ 
			{$\#OP_{ANN}$}  & 20.07G &17.63G   & 0 & 1.49G\\
			{$\#OP_{SNN}$}  & 0  & 0 &16.12G & 16.35 G \\
			{Energy ($10^{-3}$J)} &92.32 &81.10 &3.83 &10.55\\ \hline
			\shortstack{$^1$Normalized\\ Energy} &1 &0.8783& 0.0414& 0.1142\\
			\toprule[1pt]
		\end{tabular}
		\begin{tablenotes}
	\scriptsize
	\item[*] $^1$The energy consumptions of EVSNN and PA-EVSNN are 24.15$\times$ and 8.76$\times$ more computationally efficient than E2VID, respectively.
\end{tablenotes}
	\end{center}
	\vspace{-7mm}
	\label{table7}
\end{table}
\vspace{-2mm}
\section{Conclusion}

In this paper, we have presented EVSNN and PA-EVSNN models, the event-based video reconstruction models based on SNN architecture. 
We show that the spiking neurons have the capability of extracting temporal information, and SNN can achieve large scale regression tasks such as event-based video reconstruction.
Compared to E2VID, the proposed EVSNN and PA-EVSNN have 24.15$\times$ and 8.76$\times$ more computationally efficient improvement, which shows great potential of SNN for low consumption applications.
We believe that the development of energy-efficient SNN models for large-scale regression tasks is promising.

{\small
\noindent\textbf{Acknowledgments.} This work is partially supported by grants from the National Natural Science Foundation of China under contract No. 62027804, No. 61825101, and No. 62088102.}

{\small
\bibliographystyle{ieee_fullname}
\bibliography{egbib}
}

\end{document}